\documentclass{article}

\usepackage{microtype}
\usepackage{graphicx}
\usepackage{subfigure}
\usepackage{booktabs} 
\usepackage{adjustbox}
\usepackage{multirow}

\usepackage[breaklinks=true]{hyperref}

\usepackage{mathtools}
\usepackage{amsfonts}

\newcommand{\bpi}{\boldsymbol{\pi}}
\newcommand{\bbeta}{\boldsymbol{\beta}}

\newcommand{\bgamma}{\boldsymbol{\gamma}}

\DeclarePairedDelimiterX{\norm}[1]{\lVert}{\rVert}{#1}

\usepackage{color}

\usepackage[accepted]{icml2019}

\icmltitlerunning{Training CNNs with Selective Allocation of Channels}

\begin{document}

\twocolumn[
\icmltitle{Training CNNs with Selective Allocation of Channels}



\icmlsetsymbol{equal}{*}

\begin{icmlauthorlist}
\icmlauthor{Jongheon Jeong}{ee}
\icmlauthor{Jinwoo Shin}{ee,ai,aitrics}
\end{icmlauthorlist}

\icmlaffiliation{ee}{School of Electrical Engineering, KAIST, Daejeon, South Korea}
\icmlaffiliation{ai}{Graduate School of AI, KAIST, Daejeon, South Korea}
\icmlaffiliation{aitrics}{AITRICS, Seoul, South Korea}

\icmlcorrespondingauthor{Jinwoo Shin}{jinwoos@kaist.ac.kr}

\icmlkeywords{Machine Learning, ICML}

\vskip 0.3in
]



\printAffiliationsAndNotice{}  

\begin{abstract}
Recent progress in deep convolutional neural networks (CNNs) have enabled a simple paradigm of architecture design: \emph{larger models typically achieve better accuracy}. Due to this, in modern CNN architectures, it becomes more important to design models that generalize well under certain resource constraints, {e.g.\ the number of parameters.} In this paper, we propose a simple way to improve the capacity 
of any CNN model having large-scale features, without
adding more parameters.
In particular, we modify a standard convolutional layer to have a new functionality of \emph{channel-selectivity}, so that the layer is trained to select important channels to re-distribute their parameters. 
Our experimental results under various CNN architectures and datasets
demonstrate that the proposed new convolutional layer allows new optima that generalize better via efficient resource utilization, compared to the baseline.
\end{abstract}

\section{Introduction}
\vspace{-0.02in}

Convolutional neural networks (CNNs) have become one of the most effective approaches for various tasks of machine learning.
With a growing interest, there has been a lot of works on designing advanced CNN architectures \citep{cvpr/inception, corr/vggnet, icml/batchnorm, cvpr/resnet}.
Although modern CNNs are capable to scale over a thousand of layers \citep{eccv/identity} or channels \citep{cvpr/densenet}, 
deploying them in the real-world becomes increasingly difficult due to computing resource constraints.
This has motivated the recent literature such as
resource-efficient architectures \citep{cvpr/condensenet, cvpr/mobilenetv2, eccv/shuffle_v2}, 
low-rank factorization \citep{bmvc/lowrank, nips/tensorizing}, 
weight quantization \citep{eccv/xnornet, corr/binarynet, eccv/clipq} and
anytime/adaptive networks \citep{cvpr/spatial_adaptive, icml/adaptive_nn, iclr/msdnet}.

In order to design a resource-efficient CNN architecture, it is important to process succinct representations of large-scale features. 
At this point of view, there have been continuous attempts to find an efficient layer for handling such extremely large number of features \citep{corr/squeezenet, cvpr/deeproots, corr/igcv3, cvpr/mobilenetv2, eccv/shuffle_v2}. However, most prior works assume that the layer is \emph{static}, i.e., the structure in weight connectivity is unchanged during training. 
Such static layers inevitably have to allocate too many parameters across {\em homogeneous} features, since it is hard to get prior knowledge on the features before training the network. 
For instance, one of state-of-the-art models, DenseNet-BC-190 \cite{cvpr/densenet}, devotes $70\%$ of the parameters for just performing dimensionality reduction of pointwise convolutional layers.
Such an architectural inefficiency may harm the generalization ability of the model, given a fixed number of parameters.

To alleviate the issue of inefficient allocation of parameters, one can attempt to utilize the posterior information after training, e.g.\ network pruning \citep{corr/deepcomp, cvpr/he_channel, iccv/slimming}, or neural architecture search \citep{cvpr/nasnet, corr/amoebanet, nips/naonet}. A shortcoming of this direction, however, is that it typically requires a time-consuming repetition of training cycles.

\textbf{Contribution.} In this paper, we propose a new way of training CNNs so that each convolutional layer can select channels of importance dynamically during training. As the training progresses, some input channels of a convolutional layer may have almost no contribution to the output, wasting the resources allocated to the channels for the rest of the training. 
Our method detects such channels, and re-distribute the resources from those channels to another top-$K$ selected channels of importance. 
Consequently, our training scheme is a process that increases the efficiency of CNN by dynamically pruning or re-wiring its parameters on-the-fly along with learning them. 
In a sense, our method ``imitates'' how hippocampus in brain learns, where new neurons are generated and rewired daily under maintenance via neuronal apoptosis or pruning \citep{nature/increasing, nature/pattern}.

Our CNN-training method consists of two building blocks.
First, we propose the \emph{expected channel damage matrix} (ECDM), which estimates the changes of the output vector given each channel is damaged (or removed). This provides a safe criterion for selecting channels to remove (or to emphasize) during training. 
Second, we impose \emph{spatial shifting bias} for effective recycling of parameters. It turns out this allows a convolutional layer to ``enlarge'' the convolutional kernel selectively to important channels only. 

We evaluate our method on CIFAR-10/100, Fashion-MNIST, Tiny-ImageNet, and ImageNet classification datasets with a wide range of recent CNN architectures,
including ResNet \citep{cvpr/resnet} and DenseNet \citep{cvpr/densenet}. 
Despite of its simplicity, our experimental results show that training with channel-selectivity consistently improves accuracy over its counterpart across all tested architectures.  
For example, the proposed selective convolutional layer applied to DenseNet-40 provides {8.01\% relative reduction} in test error rates for CIFAR-10.  
Next, we show that our method can also be used for model compression. By applying our method to a highly-efficient CondenseNet \citep{cvpr/condensenet}, we could further improve its efficiency: the resulting model has 25$\times$ fewer FLOPs compared to ResNeXt-29 \cite{cvpr/resnext}, while achieving better accuracy.  

Compared to the significant interests on pruning parameters during training, i.e.,\ network sparsity learning \citep{nips/ssl, icml/vd_sparse, nips/sbp, nips/bcdeep, iclr/l0, icml/info_bottle}, 
the progress is arguably slower on \emph{re-wiring} the pruned parameters to maximize its utility. 
\citet{corr/dsd} proposed Dense-Sparse-Dense (DSD) training flow, showing that re-training after re-initialization of the pruned connections can further improve accuracy. Dynamic network surgery \citep{nips/dyn_surgery} introduced a method of splicing the pruned connections to recover the possibly mis-pruned ones, showing better compression performances. The recently proposed MorphNet \cite{cvpr/morphnet} attempts to find an optimal widths of each layer from shirinking and expanding a given DNN through iterative training passes. 

Our approach proposes a new way of re-wiring, with several advantages 
over the existing methods: (a) \emph{generic, easy-to-use}: it can be applied to train any kind of CNN, (b) \emph{single-pass}: it does not require any post-processing or re-training 
as it is seamlessly integrated into existing training schemes, and (c) \emph{flexibility}: it allows to easily balance between accuracy improvement and model compression on-demand. We believe our work provides a new direction on the important problem of training CNNs more efficiently.

\vspace{-0.05in}
\section{Selective Convolution}
\vspace{-0.02in}

\begin{figure}
	\centering
	\includegraphics[height=1.2in]{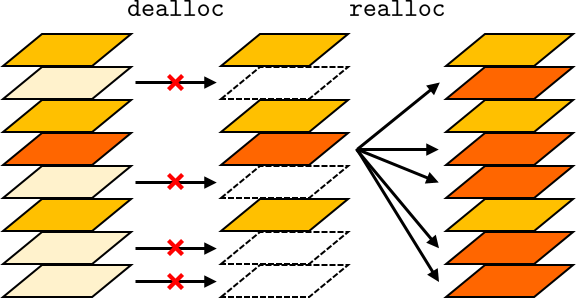}
	\caption{Illustration of channel de-allocation and re-allocation procedures. The higher the saturation of the channel color, the higher the channel importance.}
	\label{fig:ops}
	\vspace{-0.15in}
\end{figure}

Our goal is to design a new convolutional layer which can replace any existing one, with improved utilization of network parameters via selecting channels of importance.
We call the proposed layer \emph{selective convolution}.
We train this layer via two operations that make a \emph{re-distribution} of the given input channels:
\begin{enumerate}
	\vspace{-0.05in}
	\item \emph{Channel de-allocation} (\texttt{dealloc}): Obstruct unnecessary channels from being used in future computations, and release the corresponding parameters.
	\item \emph{Channel re-allocation} (\texttt{realloc}): Overwrite top-$K$ important channels into the obstructed areas, and recycle the parameters in there.
	\vspace{-0.05in}
\end{enumerate}
Figure~\ref{fig:ops} illustrates the two basic operations. More details of \texttt{dealloc} and \texttt{realloc} are described in Section~\ref{ss:comp_scu}.

During training a neural network with selective convolutional layers, 
the channel-selectivity is obtained by simply calling \texttt{dealloc} and \texttt{realloc} for each chosen layer on demand along with the standard stochastic gradient descent (SGD) methods. 
Repeating \texttt{dealloc} and \texttt{realloc} alternatively translates the original input to what has only a few important channels, potentially duplicated multiple times. 
Namely, the parameters originally allocated to handle the entire input now operate on its important subset. 

We aim to design \texttt{dealloc} and \texttt{realloc} to be \emph{function-preserving}, i.e.\ they do not change the output of the convolution. This allows us to call them anytime during SGD training without damaging the network output. On the other hand, since the resource released from \texttt{dealloc} is limited, it is also important for \texttt{realloc} to choose channels that will maximize resource utilization. 
This motivates us to design for those operations a more delicate metric of channel importance than other existing magnitude-based metrics, e.g., weight $\ell^2$-norm \cite{corr/LiKDSG16}. To this end, we propose \emph{expected channel damage matrix} (ECDM) in Section~\ref{ss:ecds}, which leads to an efficient and safe way of identifying channels with low contribution to the output.
We provide the architectural description of selective convolution
in Section~\ref{ss:comp_scu}, 
and the detailed training scheme using ECDM in Section~\ref{ss:training_inference}.

\begin{figure}
	\centering
	\includegraphics[height=0.9in]{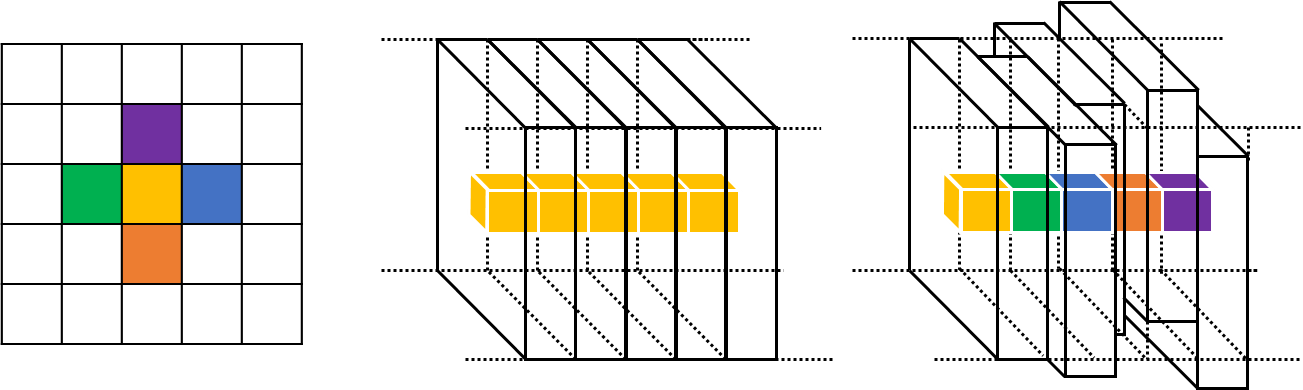}
	\vspace{-0.1in}
	\caption{The kernel enlarging effect of spatial shifting. The colored strides indicate a single patch of computation in a convolutional layer assuming the kernel size is 1 for simplicity.}
	\label{fig:spatial_bias}
	\vspace{-0.1in}
\end{figure}

\vspace{-0.05in}
\subsection{Expected Channel Damage Matrix (ECDM)}
\label{ss:ecds}
\vspace{-0.02in}

To begin with, we let $\mathrm{Conv}(\mathbf{X}; \mathbf{W})$ to denote
a convolutional layer (or function) 
for its weight $\mathbf{W}\in\mathbb{R}^{I \times O \times K^2}$ and its input random variable $\mathbf{X}\in \mathbb{R}^{I \times H \times W}$. Here, $I$ and $O$ denote the number of input and output channels, respectively, $H$ and $W$ are the height and width of the input, and $K$ denotes the kernel size.

\emph{Expected channel damage matrix} (ECDM) is designed for measuring the expected functional difference 
\begin{equation*}
\mathbb{E}_{\mathbf{X}}[\mathrm{Conv}(\mathbf{X}; \mathbf{W})-\mathrm{Conv}(\mathbf{X}; \mathbf{W}_{-i})],  
\end{equation*}
where $\mathbf{W}_{-i}$ is identical to $\mathbf{W}$ but $\mathbf{W}_{i,:,:}$ is set to 0. In other words, it measures the expected amount of changes in output when $i$-th channel is ``damaged'' or ``pruned''. 
Remark that this quantity is directly related to the function-preserving property we want to achieve.
For $i=1,\ldots, I$, we define $\mathrm{ECDM}(\mathbf{W}; \mathbf{X})_i$ by averaging the expectation over the spatial dimensions:
\begin{align*}
&\mathrm{ECDM}(\mathbf{W}; \mathbf{X})_i \in \mathbb{R}^{O} \\ 
&\coloneqq \frac{1}{HW} \sum_{h,w}\mathbb{E}_{\mathbf{X}}[ \mathrm{Conv}(\mathbf{X}; \mathbf{W}) -\mathrm{Conv}(\mathbf{X}; \mathbf{W}_{-i})]_{:, h, w}.
\end{align*}
Notice that the above definition requires a marginalization over $\mathbf{X}$. One can estimate it via Monte Carlo sampling using training data, but it can be computationally too expensive 
if it is used repeatedly during training.
Instead, we propose a simple approximation of ECDM utilizing batch normalization (BN) layer \citep{icml/batchnorm} to infer the current input distribution at any time of training, in what follows.

Consider a hidden neuron $x$ following BN and ReLU non-linearity \cite{icml/relu}, i.e.\ $y = \mathrm{ReLU}(\mathrm{BN}(x))$, and suppose one wants to estimate $\mathbb{E}[y]$ \emph{without} sampling. 
To this end, we exploit the fact that BN already ``accumulates'' its input statistics continuously throughout training. 
If we simply assume that $\mathrm{BN}(x) \sim \mathcal{N}(\beta, \gamma^2)$ (i.e.\ normal distribution),
where $\gamma$ and $\beta$ are the scaling and shifting parameter of BN, respectively, 
it is elementary to check:
\begin{equation}
\label{eq:brelu}
\mathbb{E}[y] = \mathbb{E}[\mathrm{ReLU}(\mathrm{BN}(x))] =
|\gamma| \phi_{\mathcal{N}}{\left(\frac{\beta}{|\gamma|}\right)} + \beta \Phi_{\mathcal{N}}{\left(\frac{\beta}{|\gamma|}\right)},
\end{equation}
where $\phi_{\mathcal{N}}$ and $\Phi_{\mathcal{N}}$ denote the p.d.f.\ and the c.d.f.\ of the standard normal distribution, respectively.

The idea is directly extended to obtain a closed form approximation of $\mathrm{ECDM}(\mathbf{W};\mathbf{X})$ when $\mathbf{X}$ is from $\mathrm{ReLU}(\mathrm{BN}(\cdot))$. In practice, this assumption is quite reasonable as many of nowadays CNNs adopt this $\mathrm{BN}\rightarrow\mathrm{ReLU}\rightarrow\mathrm{Conv}$ as a building block of designing a model \citep{cvpr/resnet, cvpr/densenet, cvpr/resnext, nips/dualpath}. {Under assuming that $\mathbf{X}_{i,h,w}\sim\max(\mathcal{N}(\beta_i, \gamma_i^2), 0)$ for all $i, h, w$, we obtain that
	for $i=1, \ldots, I$:}
\begin{multline}
\label{eq:ecdm_approx}
\mathrm{ECDM}(\mathbf{W}; \mathbf{X})_i \\
=
\underbrace{
	\left(|\gamma_i| \phi_{\mathcal{N}}{\left(\frac{\beta_i}{|\gamma_i|}\right)} + \beta_i \Phi_{\mathcal{N}}{\left(\frac{\beta_i}{|\gamma_i|}\right)}\right)
}_\text{(a)}
\cdot
\underbrace{
	{\sum_{k=1}^{K^2}\mathbf{W}_{i, :, k}},
}_\text{(b)}
\end{multline}
where the above equality follows from the linearity of convolutional layer, the linearity of expectation, and \eqref{eq:brelu}.
The detailed derivation is given in the supplementary material.

There are two main terms in \eqref{eq:ecdm_approx}:
(a) measures {the overall activity level of the $i$-th channel} from BN statistics, and (b) does the sum of weights related to the channel. Therefore, it allows a way to capture not only ``low-magnitude'' channels, but also channels of ``\emph{low-contribution}'' under the distribution of $\mathbf{X}$.
On the other hand, existing other magnitude-based metrics \citep{corr/LiKDSG16, iccv/slimming, nips/sbp}
typically aim only for the former.

\vspace{-0.05in}
\subsection{Selective Convolutional Layer}
\label{ss:comp_scu}
\vspace{-0.02in}

Any CNN model can have the de/re-allocation mechanism in its training,
simply by replacing each convolutional layer $\mathrm{Conv}(\mathbf{X}; \mathbf{W})$
with the proposed \emph{selective convolutional layer} $\mathrm{SelectConv}(\mathbf{X}; \mathbf{W})$. 
Compared to the standard convolution, $\mathrm{SelectConv}$ has an additional layer that rebuilds an input in channel-wise:
\begin{equation*}
\mathrm{SelectConv}(\mathbf{X};\mathbf{W}) \coloneqq \mathrm{Conv}(\mathrm{SelectChannel}(\mathbf{X}); \mathbf{W}).
\end{equation*}
In essence, $\mathrm{SelectChannel}$ requires to perform channel blocking and re-indexing for \texttt{dealloc} and \texttt{realloc}, respectively. 
One can implement this layer by:
\begin{equation}
\label{eq:selectch1}
\mathrm{SelectChannel}(\mathbf{X};\mathbf{g}, \bpi)_i \coloneqq g_i \cdot \mathbf{X}_{\pi_i},
\end{equation}
for indices $\pi_i \in \{1, 2, \ldots, I\}$ and gate variables $g_i \in \{0, 1\}$ for $i=1,\ldots, I$.
Here, multiple $\pi_i$'s can be the same, i.e.\ a channel is copied multiple times,
and $g_i=0$ means the input channel $\mathbf{X}_{\pi_i}$ is blocked.
In the case that a channel is copied $N$ times, the convolution will process the channel with $N$ times more parameters compared to the standard processing. 
However, na\"ively copying a channel in \eqref{eq:selectch1} does not give any benefit of using more parameters, due to the linearity of convolutional layer: if two input channels are identical, the corresponding weights are degenerated.
To address this issue,
we impose \emph{spatial shifting biases} $b_i = (b_i^h, b_i^w) \in \mathbb{R}^2$ for re-allocated channels:
we re-define $\mathrm{SelectChannel}$ as
\begin{equation*}
\mathrm{SelectChannel}(\mathbf{X}; \mathbf{g}, \bpi, \mathbf{b})_i \coloneqq g_i \cdot \mathrm{shift}(\mathbf{X}_{\pi_i}, b_i),
\end{equation*}
where $\mathrm{shift}(\mathbf{X}, b)$ denotes the spatial shifting operation on $\mathbf{X}$. For each pixel $(x,y)$, 
we define $\mathrm{shift}(\mathbf{X}, b)_{x,y}$ as:
\begin{align*}
\mathrm{shift}\left(\mathbf{X}, b\right)_{x,y}  \coloneqq &\sum_{n=1}^H\sum_{m=1}^W \mathbf{X}_{n,m} \nonumber\\
& \times\max\left(0, 1-|x-n+b^h|\right) \nonumber\\ 
& \times\max\left(0, 1-|y-m+b^w|\right)
\end{align*}
using a bilinear interpolation kernel. This formulation allows $b$ to be continuous real values, thereby to be learned via SGD with other parameters jointly.
{We remark that similar spatial shifting operations have recently gained attention in the area of CNN architecture design \cite{cvpr/active, iccv/deform, cvpr/shift, nips/asl}, with their efficient implementations.}
This trick encourages to utilize the re-allocated parameters effectively, as it provides diversity on the copied channels when the convolution is applied. Essentially, as illustrated in Figure~\ref{fig:spatial_bias}, it provides an effect of \emph{enlarging} the convolutional kernel, in particular, for the re-allocated channels only. In other words, our method recycles its parameters by \emph{selectively} expanding the kernel of important channels.

\vspace{-0.05in}
\subsection{Training Scheme: Channel De/Re-allocation}
\label{ss:training_inference}
\vspace{-0.02in}

Given a selective convolutional layer 
with parameters $\mathbf{S}=(\mathbf{W}, \mathbf{g}, \bpi, \mathbf{b})$, 
we design \texttt{dealloc} and \texttt{realloc} to train $\mathbf{S}$.
For example, once some channels are chosen to be de-allocated, the actual operation can be done by just setting $g_i=0$ for the channels.
We utilize ECDM in order to identify channels to be de/re-allocated.
Given a desired damage level $\gamma>0$, the objective of \texttt{dealloc} can be written as the following optimization problem:
\begin{equation*}
\begin{aligned}
& \underset{\mathbf{g}}{\text{minimize}}
& & \sum_{i=1}^I g_i \\
& \text{subject to}
& & \norm[\bigg]{\sum_{i=1}^{I} (1-g_i) \cdot \mathrm{ECDM}(\mathbf{W};\mathbf{X})_i}_{\infty} \leq \gamma, \\
&&& g_i\in \{0, 1\}, \; i = 1, \ldots, I.
\end{aligned}
\end{equation*}
However, the above combinatorial optimization is computationally intractable (i.e., NP-hard) in general 
as it is reduced to the \emph{0-1 multi-dimensional knapsack problem} (MKP) \cite{book/knapsack}. 
Although many heuristics for MKP \cite{ejor/improved_knapsack, ec/evol_knapsack} can be used for \texttt{dealloc}, 
we consider a simple greedy algorithm. First, we normalize ECDM with respect to the output dimension, namely \emph{normalized-ECDM}\footnote{In practice, using $\mathrm{nECDM}$ makes the hyperparameter $\gamma$ to be less sensitive on $I$, since $\mathrm{nECDM}(\mathbf{W}; \mathbf{X})_i$ represents \emph{relative} contributions across the input channels.} (nECDM): 
\begin{equation*}
\mathrm{nECDM}(\mathbf{W}; \mathbf{X})_{:, j} \coloneqq \frac{|\mathrm{ECDM}(\mathbf{W}; \mathbf{X})_{:, j}|}{\sum_{i=1}^I |\mathrm{ECDM}(\mathbf{W}; \mathbf{X})_{i, j}|},
\end{equation*}
for $j=1,\ldots,O$. Once nECDM is computed, channels to be de-allocated are determined by the channel of minimum $||\mathrm{nECDM}(\mathbf{W};\mathbf{X})_i||_{\infty}$ iteratively, 
while the $\ell^\infty$-norm of their vector sum of $\mathrm{nECDM}(\mathbf{W};\mathbf{X})_i$ is less than $\gamma$.

\begin{algorithm}[tb]
	\caption{Channel de-allocation (\texttt{dealloc})}
	\label{alg:dealloc}
	\begin{algorithmic}
		\STATE {\bfseries Input:} $\mathbf{S}=(\mathbf{W}, \mathbf{g}, \bpi, \mathbf{b})$, $\mathrm{nECDM}(\mathbf{W}; \mathbf{X})\in\mathbb{R}^{I\times O}$, damage level $\gamma>0$
		\vspace{0.05in}
		\hrule
		\vspace{0.05in}
		\STATE Initialize $C, C' \leftarrow \emptyset, \emptyset$
		\REPEAT
		\STATE $C \leftarrow C'$
		\STATE $C' \leftarrow C\cup \{\mathrm{argmin}_{i\not\in C} \norm{\mathrm{nECDM}(\mathbf{W};\mathbf{X})_i}_{\infty}\}$
		\UNTIL{$\norm{\sum_{i\in C'}\mathrm{nECDM}(\mathbf{W};\mathbf{X})_i}_{\infty} \leq \gamma$}
		\FORALL{$i\in C$}
		\STATE $g_i \leftarrow 0$
		\ENDFOR
	\end{algorithmic}
\end{algorithm}

\begin{algorithm}[tb]
	\caption{Channel re-allocation (\texttt{realloc})}
	\label{alg:realloc}
	\begin{algorithmic}
		\STATE {\bfseries Input:} $\mathbf{S}=(\mathbf{W}, \mathbf{g}, \bpi, \mathbf{b})$, $\mathrm{nECDM}(\mathbf{W}; \mathbf{X})\in\mathbb{R}^{I\times O}$, candidate size $K$, maximum re-allocation size $N_\mathtt{max}$
		\vspace{0.05in}
		\hrule
		\vspace{0.05in}
		\STATE Initialize $C \leftarrow \emptyset$
		\FOR{$i=1$ {\bfseries to} $I$}
		\STATE $s_i \leftarrow \norm{\mathrm{nECDM}(\mathbf{W};\mathbf{X})_i}_2$
		\STATE $N \leftarrow$ $|\{j\in\{1,\ldots,I\}:\pi_j=\pi_i\}|$
		\IF{$N > N_{\mathtt{max}}$}
		\STATE $s_i \leftarrow 0$
		\ENDIF
		\ENDFOR
		\STATE $C \leftarrow$ Select top-$K$ indices from $\mathbf{s}$
		\FOR{$i=1$ {\bfseries to} $I$}
		\IF{$g_i=0$}
		\STATE $c \leftarrow$ Select an element from $C$ randomly
		\STATE $\pi_i, g_i, \mathbf{W}_{i,:,:} \leftarrow \pi_c, 1, \boldsymbol{0}$
		\STATE Re-initialize $b_i$ randomly
		\ENDIF
		\ENDFOR
	\end{algorithmic}
\end{algorithm}

In case of \texttt{realloc}, on the other hand, we select top-$K$ {largest} channels with respect to the $\ell^2$-norm of nECDM, i.e.\ $||\mathrm{nECDM}(\mathbf{W};\mathbf{X})_{i}||_2$.\footnote{{We use $\ell^2$-norm for \texttt{realloc} to consider features that contribute across many filters more importantly. Nevertheless, we found using $\ell^\infty$-norm instead also achieves a comparable result.}} The selected top-$K$ channels randomly occupy the channels that are currently de-allocated (i.e., $g_i=0$).
When $i$-th channel is re-allocated, 
$\mathbf{W}_{i, :, :}$ are set to zero so that the operation does not harm the training.
We also set a maximum reallocation size $N_{\mathtt{max}}$ to prevent a feature to be re-allocated too redundantly.
Algorithm~\ref{alg:dealloc} and \ref{alg:realloc} summarize the overall procedure of \texttt{dealloc} and \texttt{realloc}, respectively. 

Finally, the training scheme of $\mathbf{S}$ is build upon any existing SGD training method, 
simply by calling \texttt{dealloc} or \texttt{realloc} additionally on demand. 
In other words, at any time during training $\mathbf{W}$ via SGD, \texttt{dealloc} and \texttt{realloc} additionally updates the remaining parameters of $\mathbf{S}$: \texttt{dealloc} for $\mathbf{g}$, and \texttt{realloc} for $\mathbf{g}, \bpi, \mathbf{b}$ and $\mathbf{W}$.

\vspace{-0.05in}
\section{Experiments}
\label{s:experiments}
\vspace{-0.02in}

\begin{table*}[ht]
	\caption{Comparison of test error rates on various classification tasks. ``SelectConv'' indicates our model from the corresponding baseline that is trained with channel-selectivity. We indicate $k$ by the growth rate of DenseNet. All the reported values and error bars are measured by computing mean and standard deviation across 3 trials upon randomly chosen seeds, respectively.}
	\label{table:improving_cifar}
	\vspace{0.05in}
	\centering
	\small
    \begin{adjustbox}{width=1\linewidth}
	\begin{tabular}{lccllll}
		\toprule
		   &     &       &    \multicolumn{4}{c}{Error rates (\%)}\\
        \cmidrule(l){4-7} 
		Model           &   Params      &    Method    &  CIFAR-10     &  CIFAR-100  &  Fashion-MNIST  & Tiny-ImageNet
		\\ \midrule
		DenseNet-40   & \multirow{2}{*}{0.21M} &  Baseline  
		& 6.62{\scriptsize $\pm$0.15}  
		& 29.9{\scriptsize $\pm$0.1} 
		& 5.03{\scriptsize $\pm$0.07}  
		& 45.8{\scriptsize $\pm$0.2}  \\ 
		(bottleneck, $k=12$) & & SelectConv 
		& \textbf{6.09{\scriptsize $\pm$0.10} (-8.01\%)}  
		& \textbf{28.8{\scriptsize $\pm$0.1} (-3.42\%)} 
		& \textbf{4.73{\scriptsize $\pm$0.06} (-5.96\%)} 
		& \textbf{44.4{\scriptsize $\pm$0.2} (-3.03\%)}
		\\ \midrule
		DenseNet-100 & \multirow{2}{*}{1.00M} & Baseline  
		&  4.51{\scriptsize $\pm$0.04} 
		&  22.8{\scriptsize $\pm$0.3} 
		&  4.70{\scriptsize $\pm$0.06} 
		&  41.0{\scriptsize $\pm$0.1} \\
		(bottleneck, $k=12$) & & SelectConv   
		&  \textbf{4.29{\scriptsize $\pm$0.08} (-4.88\%)}   
		&  \textbf{22.2{\scriptsize $\pm$0.1} (-2.64\%)} 
		&  \textbf{4.58{\scriptsize $\pm$0.05} (-2.55\%)}  
		&  \textbf{39.9{\scriptsize $\pm$0.3} (-2.78\%)}
		\\ \midrule
		{ResNet-164 } & \multirow{2}{*}{1.66M} &  Baseline   
		&  4.23{\scriptsize $\pm$0.15}     
		&  21.3{\scriptsize $\pm$0.2} 
		&  4.53{\scriptsize $\pm$0.04}     
		&  37.7{\scriptsize $\pm$0.4} \\ 
		(bottleneck, pre-act) & & SelectConv  
		&  \textbf{3.92{\scriptsize $\pm$0.14} (-7.33\%)}   
		&  \textbf{20.9{\scriptsize $\pm$0.2} (-1.97\%)}  
		&  \textbf{4.37{\scriptsize $\pm$0.03} (-3.53\%)}  
		&  \textbf{37.5{\scriptsize $\pm$0.2} (-0.56\%)}  
		\\ \midrule
		\multirow{2}{*}{ResNeXt-29 ($8\times64d$)} & \multirow{2}{*}{33.8M} &  Baseline  
		& 3.62{\scriptsize $\pm$0.12}    
		& 18.1{\scriptsize $\pm$0.1}   
		& 4.40{\scriptsize $\pm$0.07}     
		& 31.7{\scriptsize $\pm$0.3} \\ 
		& & SelectConv  
		& \textbf{3.39{\scriptsize $\pm$0.14} (-6.36\%)} 
		& \textbf{17.6{\scriptsize $\pm$0.1} (-2.92\%)}    
		& \textbf{4.27{\scriptsize $\pm$0.06} (-2.95\%)} 
		& \textbf{31.4{\scriptsize $\pm$0.3} (-0.88\%)} 
		\\ \bottomrule
	\end{tabular}
    \end{adjustbox}
    \vspace{-0.2in}
\end{table*}
\begin{table*}[ht]
    \caption{Comparison of performance on CIFAR-10 between different CNN models including ours. Models named ``X-Pruned'' are the results from Network slimming \cite{iccv/slimming}.}
    \label{table:designing}
    \vspace{0.05in}
    \centering
    \small
    \begin{tabular}{llll}
		\toprule
		Model    & Params & \ \ \ FLOPs  & Error rates (\%)  \\ 
		\midrule
		ResNet-1001 \cite{eccv/identity}  & 16.1M  & \ \ 2,357M  & 4.62 \\
		WideResNet-28-10 \cite{bmvc/wide} & 36.5M  & \ \ 5,248M  & 4.17 \\
		ResNeXt-29 ($16\times64d$) \cite{cvpr/resnext} & 68.1M  & 10,704M & 3.58 \\
		\midrule
		VGGNet-Pruned \cite{iccv/slimming}    &     2.30M      &    \quad\ 391M      &    6.20  \\  
		ResNet-164-Pruned \cite{iccv/slimming} &     1.10M      &    \quad\ 275M      &    5.27     \\
		DenseNet-40-Pruned \cite{iccv/slimming} &     0.35M      &    \quad\ 381M      &    5.19     \\
		\midrule
		DenseNet-BC-190 \cite{cvpr/densenet} & 25.6M  & \ \ 9,388M  & 3.46 \\
					\textbf{DenseNet-BC-SConv-190 (Ours)} & \textbf{11.5M (-55.1\%)}  & \ \  \textbf{4,287M (-54.3\%)}  & \textbf{3.45 (-0.29\%)} \\
		\midrule
					CondenseNet-182 \cite{cvpr/condensenet}  &     4.20M      &  \quad\ 513M      &    3.76     \\
		\textbf{CondenseNet-SConv-182 (Ours)} & \textbf{3.24M (-22.9\%)} & \quad\ \textbf{422M (-17.7\%)} & \textbf{3.50 (-6.91\%)} 
		\\ \bottomrule
	\end{tabular}
	\vspace{-0.15in}
\end{table*}

We evaluate our method on various image classification tasks: CIFAR-10/100 \citep{dataset/cifar}, Fashion-MNIST \citep{dataset/fmnist},
Tiny-ImageNet\footnote{\url{https://tiny-imagenet.herokuapp.com/}},
and ImageNet \citep{dataset/ilsvrc} datasets.
We consider a variety of CNN architectures recently proposed, including ResNet \citep{cvpr/resnet}, DenseNet \citep{cvpr/densenet}, and ResNeXt \citep{cvpr/resnext}.
Unless otherwise stated, we fix $\gamma=0.001$, $K=3$, and $N_{\mathtt{max}}=32$ for training selective convolutional layers.
In cases of DenseNet-40 and ResNet-164, we do not use $N_{\mathtt{max}}$, i.e.\ $N_{\mathtt{max}}=\infty$, as they handle relatively fewer channels.
We did not put much effort for the very best hyperparameters of our method, so there can be better ones depending across datasets. Nevertheless, we found our method has resilience on the given configuration, as we verify from the experiments: 
it generally yield good performances for all the tested models and datasets, even in the large-scale ImageNet experiments.
The more training details, e.g.\ datasets and model configurations, are given in the supplementary material.

In overall, our results show that training with channel-selectivity consistently improves the model efficiency, mainly demonstrated in two aspects: (a) improved accuracy and (b) model compression.
We also perform an ablation study to verify the effectiveness of our main ideas. 

\vspace{-0.05in}
\subsection{Improved Accuracy with Selective Convolution}
\label{ss:improve}
\vspace{-0.02in}

We compare classification performance of various CNN models trained with our method against conventional training. For each baseline model, we consider the counterpart \emph{selective} model that every convolutional layer is replaced by the corresponding selective convolutional layer. We train the pair of models for the same number of epochs. 

\begin{table}
    \vspace{-0.1in}
    \caption{Comparison of the single-crop top-1 validation error rates on the ImageNet dataset. ``SelectConv'' indicates our model from the corresponding baseline that is trained with channel-selectivity. In case of DenseNet-121, $\gamma$ is adjusted to 0.0005.}
	\label{table:improving_imagenet}
	\vspace{0.05in}
    \centering
    \small
    \begin{tabular}{lccc}
		\toprule
		Model &  Params &   Method     & Error (\%)  \\ \midrule
		DenseNet-121 & \multirow{2}{*}{7.98M} & Baseline    & 24.7 \\
		($k=32$) & & SelectConv    & \textbf{24.4}   \\
		\midrule
		ResNet-50 & \multirow{2}{*}{22.8M} & Baseline  &  23.9 \\ 
		(bottleneck) & & SelectConv & \textbf{23.4} \\ 
		\bottomrule
	\end{tabular}
	\vspace{-0.2in}
\end{table}

Table~\ref{table:improving_cifar} and Table~\ref{table:improving_imagenet} summarize the main results. In overall, our method consistently reduces classification error rates across all the tested models compared to the conventional training.\footnote{We remark that the reduction in the ImageNet results (Table~\ref{table:improving_imagenet}) is quite non-trivial, e.g.\ reducing error $23.6\% \rightarrow 23.0\%$ requires to add 51 more layers from ResNet-101 (i.e., ResNet-152), according to the official repository: \url{https://github.com/KaimingHe/deep-residual-networks}.}
As the selective models have almost the same
number of parameters with the baseline model, the results confirm that the proposed channel de/re-allocation scheme utilized the given parameters more efficiently, i.e.\ by enlarging the kernel size of each important channel selectively. 

Recall that our training method is compatible on any existing training scheme, since \texttt{dealloc} and \texttt{realloc} does not affect the loss during training due to their function-preserving property. The training configurations used in our experiments, e.g.\ weight decay or momentum, are one of the most common choice for training CNNs. Even though not explored in this paper, we believe that the effectiveness of our method can be further improved by using more coarse-grained training schemes, e.g.\ channel-level regularization \citep{nips/ssl, iccv/slimming, nips/sbp}, as our method operates in the channel-level as well.

\vspace{-0.05in}
\subsection{Model Compression with Selective Convolution}
\vspace{-0.02in}

Next, we show that our method can also be used for model compression, when the model is under regime of large-scale features. To this end, we consider two state-of-the-art CNN models, namely DenseNet-BC-190 \cite{cvpr/densenet} and CondenseNet-182 \cite{cvpr/condensenet}. Here, CondenseNet is a highly-efficient mobile-targeted CNN architecture, outperforming MobileNet \cite{corr/mobilenet} and ShuffleNet \cite{cvpr/shufflenet}. 
We select these two architectures to compare since they both attempt to design an efficient architecture under extremely large number of features.
Here, we apply our ECDM-based channel de-allocation scheme upon these architectures to show that our method can further exploit the inefficiency of large-scale feature regime to improve the model efficiency. 
Namely, we compare the model efficiency of the models with our counterpart models with selective convolution trained using only \texttt{dealloc}, with the intention of maximizing the computational efficiency.

In the case of CondenseNet,
the architecture contains a channel pruning mechanism during training, namely \emph{learned group convolution} (LGC) layer, which is similar to \texttt{dealloc} in our method. We aim to compare this mechanism with selective convolution trained using only \texttt{dealloc}, showing that our de-allocation mechanism with ECDM can further improve the efficiency.\footnote{For the interested readers, we present the more details of LGC in the supplementary material.}
To this end, we consider a variant of CondenseNet-182 where only each of the LGC layers inside is replaced by the selective convolutional layer, coined \emph{CondenseNet-SConv-182}. We remark that, unlike LGC, we use neither group convolution nor group-lasso regularization for selective convolutional layers, {even if they can further improve the efficiency}. The other training details are set identical to the original one by {\citet{cvpr/condensenet}} for fair comparison.

Table~\ref{table:designing} report the result. First, observe that CondenseNet-SConv-182 shows much better efficiency compared to the original CondenseNet-182. Namely, our model achieves -22.9\% of reduction in parameters, and -6.91\% in error rates over the baseline, even the orignal CondenseNet-182 also performs channel pruning on $\frac{3}{4}$ of the input channels. This result confirms the effectiveness of our ECDM-based de-allocation scheme over the $\ell^1$-based LGC, suggesting a benefit of using our method for both accuracy improvement and model compression. DenseNet-BC-SConv-190, on the other hand, shows more reduction in parameters and FLOPs without loss of accuracy, compared to the original model. 
This reduction is due to that
DenseNet-BC-190 is less optimized in its architecture for efficiency, compared to CondenseNet-182.
This shows that our method can discover an inherited inefficiency inside the model as well. 
We also compare the result with various state-of-the-art CNN models.
Remarkably, CondenseNet-SConv-182 achieves even better accuracy than ResNeXt-29, while ours has 25$\times$ fewer FLOPs. Compared to \emph{network slimming} \cite{iccv/slimming}, on the other hand, this model shows significantly better accuracy with similar FLOPs.

\vspace{-0.05in}
\subsection{Ablation Study}
\label{ss:case}
\vspace{-0.02in}

We also conduct an ablation study on the proposed method, investigating the detailed analysis on our method. Throughout this study, we consider DenseNet-40, which consists 3 \emph{dense blocks}, each of which consists of 6 consecutive \emph{dense units}. Each of the units produces $k=12$ features, and those features are concatenated over the units. Unlike \citet{cvpr/densenet}, we do not place a feature compression layer between the dense blocks for simplicity.
All the experiments in this section are performed on CIFAR-10. 

\begin{figure}[t]
	\centering
	\includegraphics[height=1.5in]{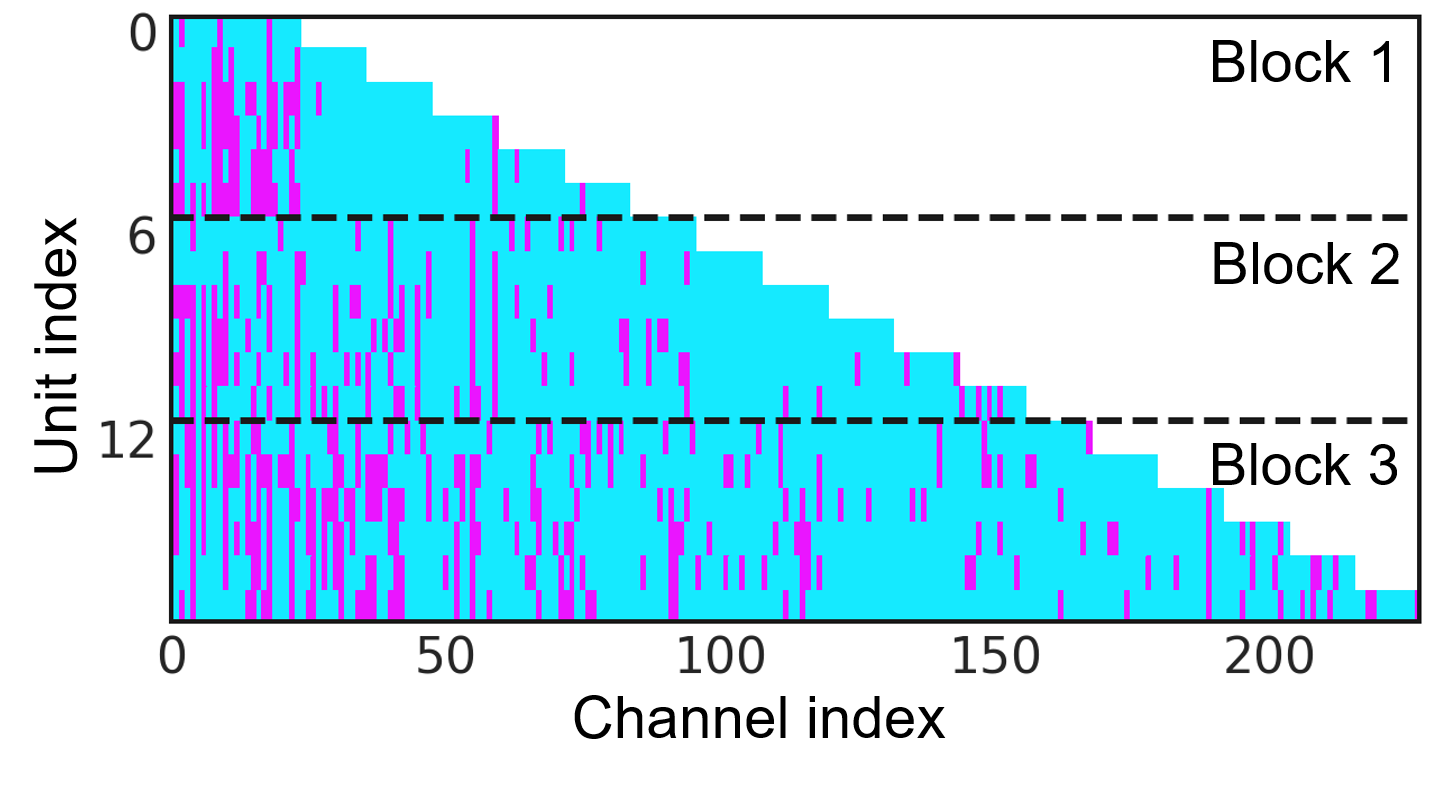}
	\vspace{-0.15in}
	\caption{Illustration of channel indices that a channel de/re-allocation is occurred in a DenseNet-40 model for each of dense units. The channels of interest are marked by magenta. Unit indices are divided into three for each dense block.}
	\label{fig:d40_2}
	\vspace{-0.1in}
\end{figure}

\textbf{Analysis on the selected channels.}
We train a DenseNet-40 model with channel selectivity, and analyze which channels are de/re-allocated during training. Figure~\ref{fig:d40_2} demonstrates channel-indices that are de/re-allocated for each dense unit. The result show that features made at early units (i.e.\ lower-level features) are de-allocated more than the others, which is consistent with our intuition. Remarkably, one can also find there are some channels which tend to be de-allocated across multiple \emph{consecutive} units, possibly across multiple blocks, but apparently used in later units. This tendency found by selective convolution reflects how DenseNet processes features under its architectural benefit: some low-level features may not needed for a long term in the visual pathway. Our method effectively utilizes the redundancy from just ``keeping'' such of the features.  

\begin{figure}[t]
	\centering
	\includegraphics[height=1.45in]{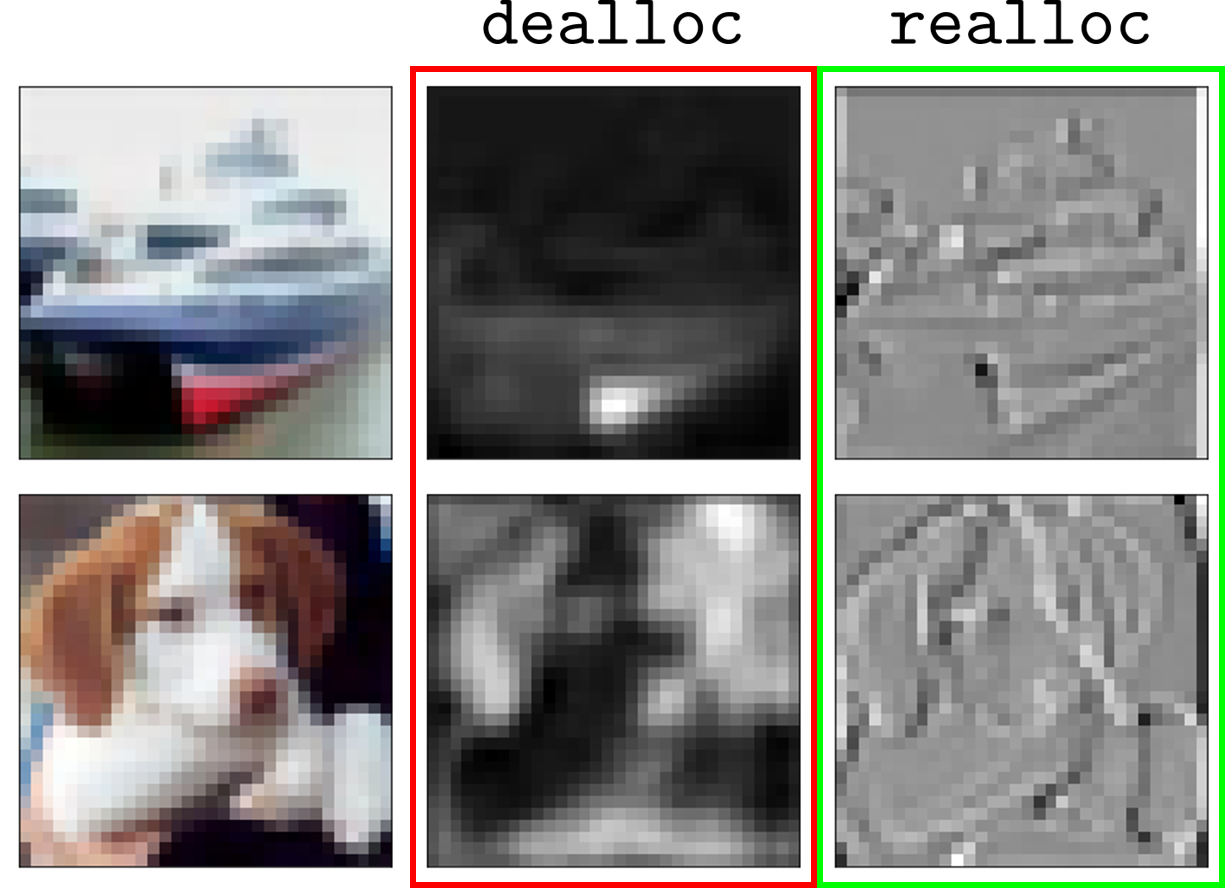}
	\vspace{-0.05in}
	\caption{Images from CIFAR-10 and their feature maps at a de-allocated channel index (middle), and at the corresponding re-allocated index (right). The feature maps are taken from the first dense unit of a DenseNet-40 model.}
	\label{fig:dr_compare}
	\vspace{-0.1in}
\end{figure}

Figure~\ref{fig:dr_compare} provides an additional insight from which channel is actually de/re-allocated in the model. {We observed that channels containing more information for the given task, e.g.\ sharp edge information for classification, tend to be re-allocated more, possibly for better processing of the task information. We provide more illustrations about which channels are de/re-allocated in the supplementary material.}

\begin{figure}[t]
	\centering
	\includegraphics[height=1.45in]{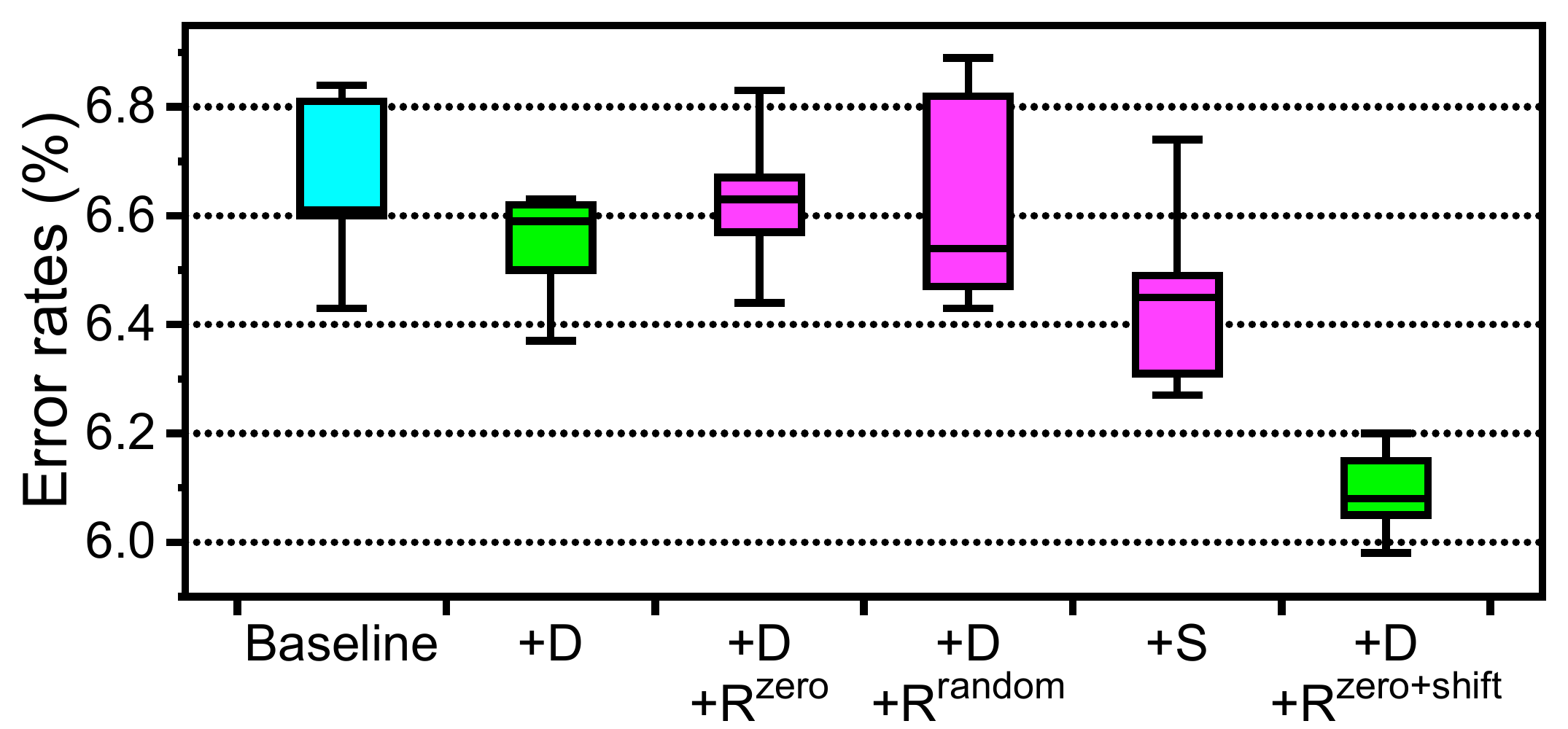}
	\vspace{-0.1in}
	\caption{Comparison of error rates on DenseNet-40 models between our model of using spatial shifting (+D+R$^{\tt zero+shift}$), and its ablations (+D+R$^{\tt zero}$ and +D+R$^{\tt random}$). Error rates of the baseline model without channel-selectivity is also provided.}
	\label{fig:shifting_box}
	\vspace{-0.1in}
\end{figure}

\begin{figure*}[t]
	\centering
	\subfigure[Non-boundary pixels]
	{
		\includegraphics[width=0.2\linewidth]{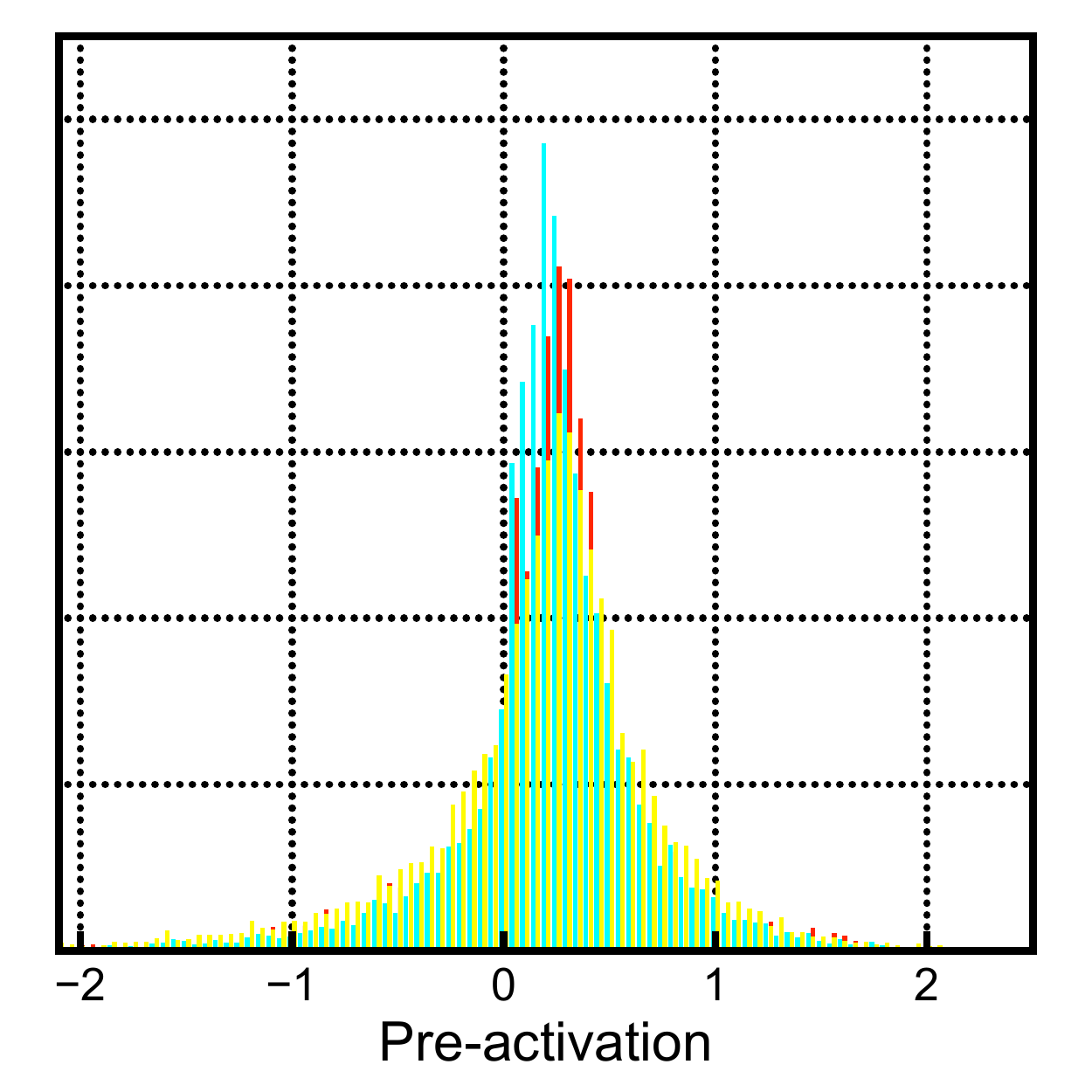}
		\label{fig:idnorm_dist_nb}
	}
	\subfigure[Boundary pixels]
	{
		\includegraphics[width=0.2\linewidth]{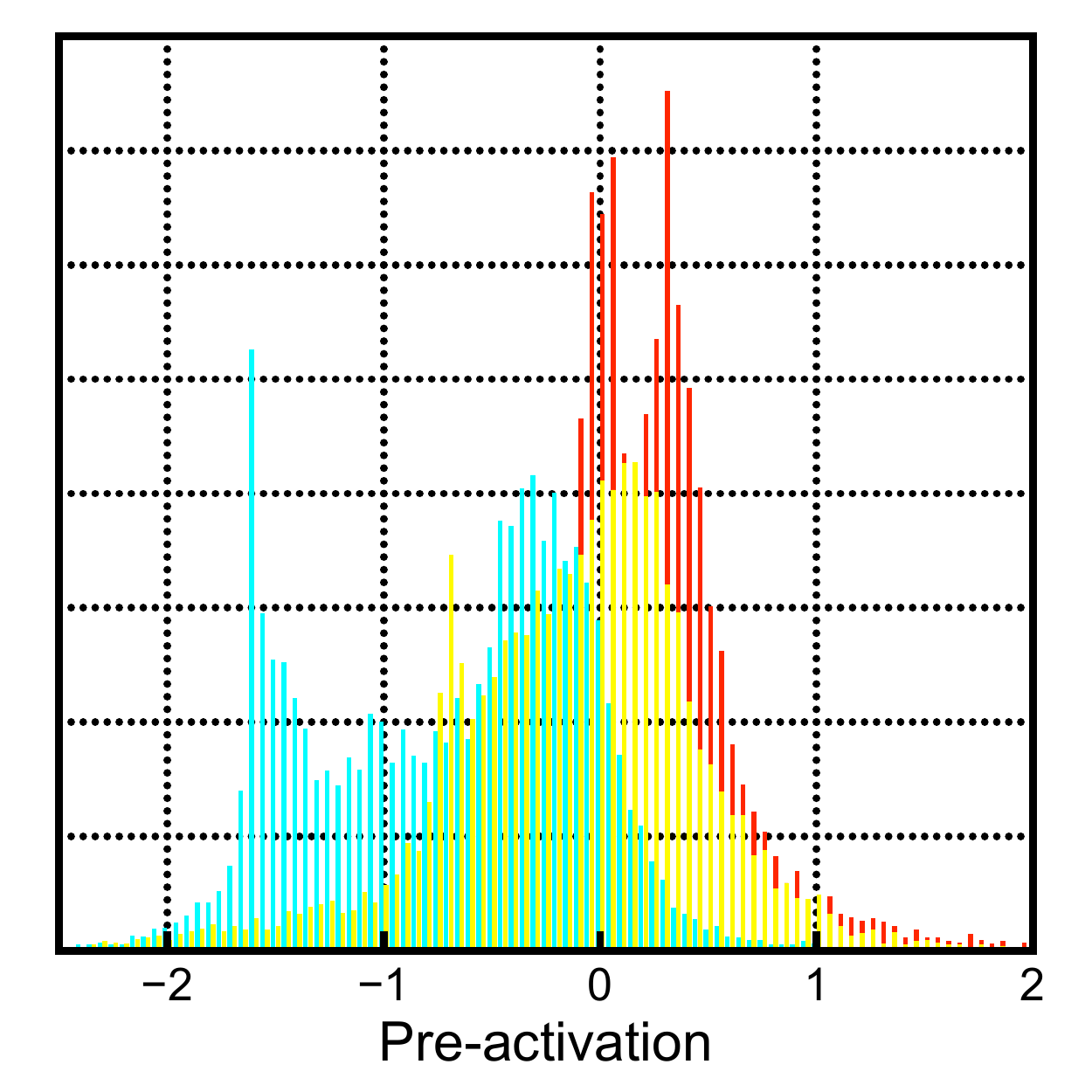}
		\label{fig:idnorm_dist_b}
	}
	\subfigure[Initialized randomly]
	{
		\includegraphics[width=0.2\linewidth]{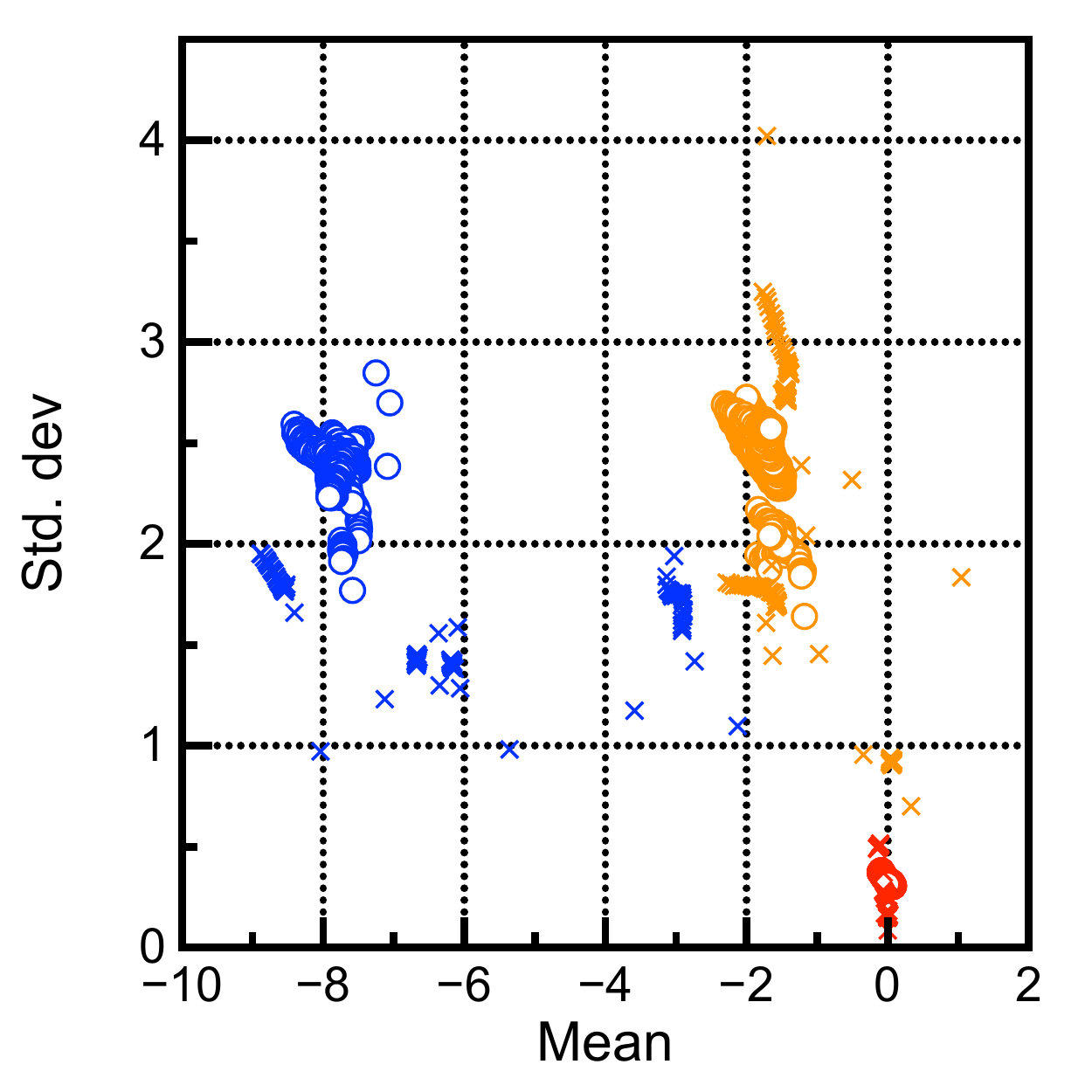}
		\label{fig:idnorm_ch_r}
	}
	\subfigure[Trained to converge]
	{
		\includegraphics[width=0.2\linewidth]{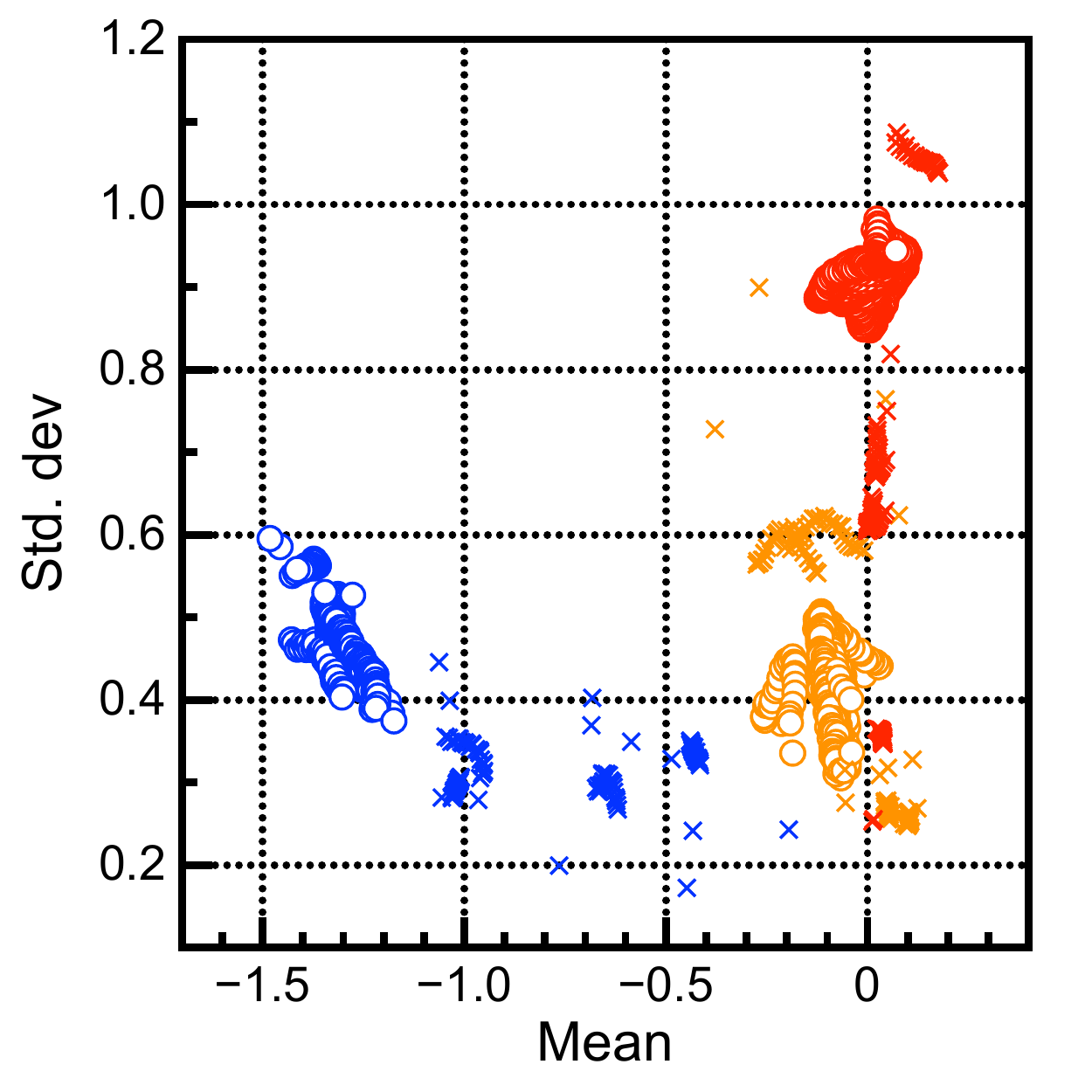}
		\label{fig:idnorm_ch}
	}
	\vspace{-0.1in}
	\caption{Pixel-wise input distributions of the $4^{\text{th}}$ layer (i.e.\ the middle of the first dense block) in a DenseNet-40 model. (a, b) Empirical distributions of three randomly chosen pixels in a fixed channel of input, which are inferred from CIFAR-10 test dataset. (c, d) Scatter plots between empirical mean and standard deviation of each pixel distributions, plotted for 3 representative channels in the input. Each plot consists 1,024 points, as a channel have $32\times 32$ pixels. Pixels in boundaries are specially marked as $\times$.}
	\label{fig:idnorm}
	\vspace{-0.1in}
\end{figure*}

\begin{figure}[t]
	\centering
	\includegraphics[height=1.5in]{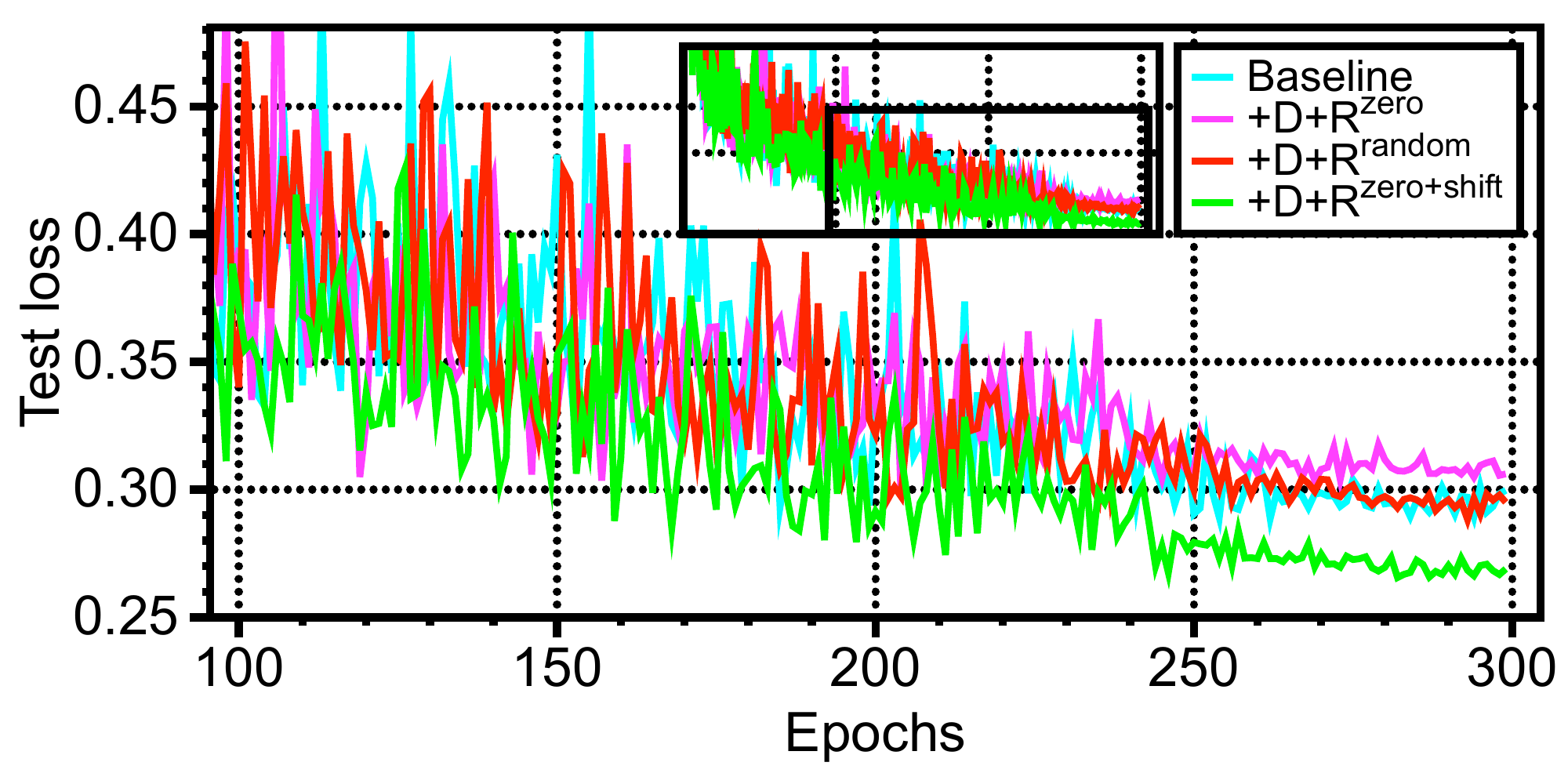}
	\vspace{-0.15in}
	\caption{Comparison of test losses of our model (+D+R$^{\tt zero+shift}$) and its ablations on spatial shifting (+D+R$^{\tt zero}$ and +D+R$^{\tt random}$). Each curve denotes cross-entropy loss measured on the CIFAR-10 test set for each training epoch.}
	\label{fig:tloss_trend}
	\vspace{-0.1in}
\end{figure}

\begin{figure}[t]
	\centering
	\includegraphics[height=1.1in]{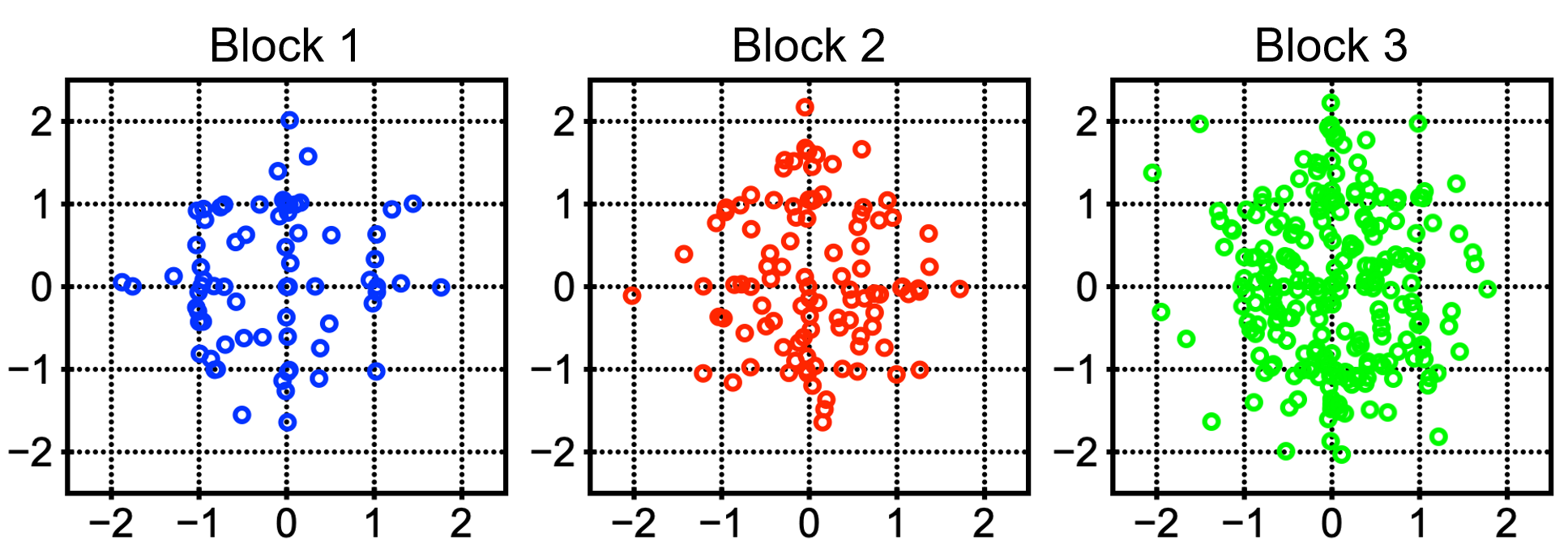}
	\vspace{-0.12in}
	\caption{Scatter plots of learned spatial biases in pixels for the re-allocated channels in a DenseNet-40 model. The points are grouped into different plots for each dense block.}
	\label{fig:d40_bias}
	\vspace{-0.12in}
\end{figure}

\textbf{Channel re-allocation.}
Our main motivation to introduce spatial shifting is
to force the \texttt{realloc} procedure to utilize the re-allocated parameters diversely in input distributions. 
To evaluate its effect in accuracy, we compare five DenseNet-40 models  
with different re-allocation scheme: 
{
	\begin{itemize}
		\item \emph{Ours} (+D+R$^{\tt zero+shift}$): If a channel is re-allocated, the corresponding convolutional weights are set to 0, and spatial shifting is imposed correspondingly. 
		\item \emph{Zero re-initialization} (+D+R$^{\tt zero}$): 
		We do not use spatial shifting from the above original configuration, i.e.,
		\texttt{realloc} is used, but spatial shifting is not imposed.
		\item \emph{Random re-initialization} (+D+R$^{\tt random}$): 
		Now, we modify the initialization, i.e.,
		\texttt{realloc} is used without spatial shifting, and the weights are re-initialized following the model initialization scheme.
		\item \emph{De-allocation only} (+D): Only \texttt{dealloc} is used, i.e.\ \texttt{realloc} is not performed during training.
		\item {\emph{Shift only} (+S): Neither de/re-allocation is used, but all channels learn spatial bias from the beginning. This ablation is essentially equivalent to the method proposed by \citet{cvpr/active}.}
	\end{itemize}
}

Figure~\ref{fig:shifting_box} clearly shows that +D+R$^{\tt zero+shift}$ outperforms the others, while +D+R$^{\tt zero}$ or +D+R$^{\tt random}$ could not statistically improve its accuracy over the baseline and +D even though \texttt{realloc} is performed. This confirms that copying a channel na\"ively is not enough,
and the spatial shifting is an effective trick
under our channel de/re-allocation setting.
In case of +S, on the other hand, we found a certain gain from the use of shifting biases, but +D+R$^{\tt zero+shift}$ also outperforms it by a large margin. 
We also emphasize that +D+R$^{\tt zero+shift}$ uses much less shifting, e.g.\ about 5 times less than +S, as it performs shifting only for the re-allocated channels. 
This confirms that our de/re-allocation scheme is crucial for the effectiveness.

Figure~\ref{fig:tloss_trend} further compares the models with the testing loss curves. Each of the curves is taken from the model that showed median performance across the trials. One can clearly observe that +D+R$^{\tt zero+shift}$ is converged at much lower testing loss, while the others are stuck at a similar local minima. Recent works show that all sub-optimal local minima in a neural network can be eliminated theoretically by adding a neuron of a certain form \cite{nips/elim_one, corr/elim_allbad}. In this sense, our training scheme can be thought as a process of continuously adding new neurons into the network during training, along with the spirit of network pruning. 

\textbf{Learned biases from spatial shifting.}
As explained in Section~\ref{ss:comp_scu}, channel re-allocation with spatial shifting has an effect of enlarging the kernel sizes for the selected channels. Figure~\ref{fig:d40_bias} illustrates how the spatial biases are actually trained for each dense block in a DenseNet-40 model with channel-selectivity.  
Interestingly, we found that some biases are converged with a tendency to \emph{align} on the pixel grids, especially at the first dense block. Since the information contained in CIFAR-10 images is given in pixel-wise, biases on exact grids will maximize the amount of new information from the original channel. The observation suggests that optimizing shifting-parameters $\mathbf{b}$ indeed considers such an effect during training. Biases at the later layers, on the other hand, shows less tendency of aligning but a larger diversity on the values, which corresponds to larger kernel sizes.

\textbf{Empirical support on the ECDM formula.}
In order to obtain the practical formula for ECDM \eqref{eq:ecdm_approx}, we assumed that the input $\mathbf{X}$ is of the from $\mathrm{ReLU}(\mathrm{BN}(\mathbf{Y}; \bbeta, \bgamma))$ for another random variable $\mathbf{Y}$, and approximated $\mathbf{X}_{i, h, w}$ by $\max(\mathcal{N}(\beta_i, \gamma_i^2), 0)$ for all $i, h, w$. Essentially, this approximation imposes two key assumptions on $\mathbf{Y}$ accordingly: (a) $\mathbf{Y}$ follows normal distribution, and (b) for a fixed $i$, each of $\mathbf{Y}_{i, :, :}$ are \emph{identically} distributed. Here, our question is that how much these assumptions hold in modern CNNs.

To validate this, 
we calculate hidden inputs $\mathbf{X}^{\texttt{test}}$ at the $4^{\text{th}}$ dense unit of a DenseNet-40 model using CIFAR-10 test images.
By analyzing empirical distributions of $\mathbf{X}^{\texttt{test}}_{c, h, w}$ for varying $h$ and $w$, we found that: (a) for a fixed $c$, most of the distributions  
are uni-modal, with exceptions at the boundary pixels (Figure~\ref{fig:idnorm_dist_nb},~\ref{fig:idnorm_dist_b}), and (b) for a large portion of $c$
the means and variances of $\mathbf{X}^{\texttt{test}}_{c, h, w}$'s are concentrated in a cluster (Figure~\ref{fig:idnorm_ch}). 
These observations support that the proposed assumptions are reasonable, with some exceptional points, e.g.\ the boundary pixels. 
We also found that the trends still exist even the model re-initialized (Figure~\ref{fig:idnorm_ch_r}), i.e.\ they are not ``learned'', but come from some structural properties of CNN. Two of such properties can be responsible: (a) the central limit theorem from the linear, weighted summing nature of convolution, and (b) equivariance of convolution on spatial dimensions. This observation confirms that ECDM is valid at anytime during training.

\vspace{-0.05in}
\section{Conclusion}
\vspace{-0.02in}

We address a new fundamental problem of training CNNs given restricted neural resources,
where our new approach is exploring pruning and re-wiring on demand via channel-selectivity.
Such a dynamic training scheme have been considered difficult
in a conventional belief, as it easily makes the training unstable.
We overcome this with our novel metric, ECDM, which allows
more robust pruning during training, consequently opens a new direction of training CNNs.
We expect that the channel-selectivity is also a desirable property for many subjects related to CNNs, e.g.\ interpretability \citep{iccv/grad_cam}, and robustness \citep{iclr/adv}, just to name a few.

\section*{Acknowledgements}

This research was supported by Naver Labs and the Engineering Research Center Program through the National Research Foundation of Korea (NRF) funded by the Korean Government MSIT (NRF-2018R1A5A1059921).

\nocite{iccv/interleaved}

\bibliography{references}
\bibliographystyle{icml2019}

\end{document}


\onecolumn
	\clearpage
	\begin{center}{\bf {\LARGE Supplementary Material}}
	\end{center}
	\begin{center}{\bf {\Large Training CNNs with Selective Allocation of Channels}}
	\end{center}
	
	\appendix
	
	\section{Detailed Derivation of ECDM Formula}\label{sec:proof}
	
	Consider a convolutional layer $\mathrm{Conv}(\mathbf{X}; \mathbf{W})$ with $\mathbf{W}\in\mathbb{R}^{I\times O\times K^2}$, and $\mathbf{X}\in \mathbb{R}^{I\times H\times W}$. 
	First, from the assumption that $\mathbf{X}_{i, h, w}\sim\max(\mathcal{N}(\beta_i, \gamma_i^2), 0)$ for all $i, h, w$, we get:
	\begin{align}
	\label{eq:expect_x}
	\mathbb{E}[\mathbf{X}_{i, h, w}] &= \int_{-\infty}^{\infty} \frac{\max(x, 0)}{|\gamma_i|\sqrt{2\pi}}\cdot\exp{\left(-\frac{(x-\beta_i)^2}{2\gamma_i^2}\right)} dx \nonumber\\
	&= \int_{0}^{\infty} \frac{x}{|\gamma_i|\sqrt{2\pi}}\cdot\exp{\left(-\frac{(x-\beta_i)^2}{2\gamma_i^2}\right)} dx \nonumber\\
	&= \int_{-\frac{\beta_i}{|\gamma_i|}}^{\infty} \frac{|\gamma_i|y + \beta_i}{\sqrt{2\pi}}\cdot\exp{\left(-\frac{y^2}{2}\right)} dy \nonumber\\
	&= \frac{|\gamma_i|}{\sqrt{2\pi}}\int_{-\frac{\beta_i}{|\gamma_i|}}^{\infty} y \cdot \exp{\left(-\frac{y^2}{2}\right)} dy 
	+ \beta_i \Phi_{\mathcal{N}}\left(\frac{\beta_i}{|\gamma_i|}\right) \nonumber\\
	&= |\gamma_i|\phi_{\mathcal{N}}\left(\frac{\beta_i}{|\gamma_i|}\right) 
	+ \beta_i \Phi_{\mathcal{N}}\left(\frac{\beta_i}{|\gamma_i|}\right) \eqqcolon f(\mathbf{X})_i,
	\end{align}
	where $\phi_{\mathcal{N}}$ and $\Phi_{\mathcal{N}}$ denote the p.d.f.\ and the c.d.f.\ of the standard normal distribution, respectively.
	
	Secondly, once noticed that $\mathrm{Conv}(\mathbf{X}; \mathbf{W}) -\mathrm{Conv}(\mathbf{X}; \mathbf{W}_{-i})$ in the definition of ECDM is identical to $\mathrm{Conv}(\mathbf{X}_{i}; \mathbf{W}_{i, :, :})$, the desired formula follows from the linearity of convolutional layer, the linearity of expectation, and \eqref{eq:expect_x}:  
	\begin{align}
	\mathrm{ECDM}(\mathbf{W}; \mathbf{X})_i &\coloneqq \frac{1}{HW} \sum_{h,w}\mathbb{E}_{\mathbf{X}}[ \mathrm{Conv}(\mathbf{X}; \mathbf{W}) -\mathrm{Conv}(\mathbf{X}; \mathbf{W}_{-i})]_{:, h, w} \nonumber\\
	&= \frac{1}{HW}\sum_{h, w}\mathbb{E}[\mathrm{Conv}(\mathbf{X}_{i}; \mathbf{W}_{i, :, :})]_{:, h, w} \nonumber\\
	&= \frac{1}{HW}\sum_{h, w}\left(\mathbb{E}\left[\sum_{x=-\floor{\frac{K}{2}}}^{\floor{\frac{K}{2}}}\sum_{y=-\floor{\frac{K}{2}}}^{\floor{\frac{K}{2}}}{\mathbf{W}_{i,j,\left(\floor{\frac{K}{2}}+x\right)\cdot K + \left(\floor{\frac{K}{2}}+y\right)}\cdot \mathbf{X}_{i, h+x, w+y}}\right]\right)_{j=1}^{O} \label{eq:ecdm_expand}\\
	&= \frac{1}{HW}\sum_{h, w}\left(\sum_{x=-\floor{\frac{K}{2}}}^{\floor{\frac{K}{2}}}\sum_{y=-\floor{\frac{K}{2}}}^{\floor{\frac{K}{2}}} \mathbf{W}_{i,j,\left(\floor{\frac{K}{2}}+x\right)\cdot K + \left(\floor{\frac{K}{2}}+y\right)}\cdot\mathbb{E}\left[\mathbf{X}_{i, h+x, w+y}\right]\right)_{j=1}^{O} \nonumber\\
	&= \frac{1}{HW}\sum_{h, w}\left(f(\mathbf{X})_i\cdot\sum_{k=1}^{K^2}\mathbf{W}_{i,j,k}\right)_{j=1}^{O}
	= f(\mathbf{X})_i \cdot \sum_{k=1}^{K^2}\mathbf{W}_{i,:,k} \nonumber\\
	&= \left(|\gamma_i|\phi_{\mathcal{N}}\left(\frac{\beta_i}{|\gamma_i|}\right) 
	+ \beta_i \Phi_{\mathcal{N}}\left(\frac{\beta_i}{|\gamma_i|}\right)\right) \cdot \sum_{k=1}^{K^2}\mathbf{W}_{i,:,k} 
	\end{align}
	Here, in \eqref{eq:ecdm_expand}, we assume that the convolutional layer uses the \emph{same padding} scheme, i.e.\ indexing of $\mathbf{X}$ outside the pixel scope is considered to be the nearest in-scope pixel.
	
	\section{Training Details}
	
	\subsection{Training Setup}
	We train every model via stochastic gradient descent (SGD) with Nesterov momentum of weight 0.9 without dampening. We use a cosine shape learning rate schedule \citep{corr/sgdr} which starts from $0.1$ and decreases gradually to $0$ throughout the training. We set a weight decay of $10^{-4}$, except for the spatial shifting biases $\mathbf{b}$ in which $10^{-5}$ is used instead.
	During the training, we call \texttt{dealloc} and \texttt{realloc} at each epoch for the half of the total epochs. 
	This is mainly for two reasons: (a) usually, most of \texttt{dealloc} is done before that time, and (b) we found this makes the training less sensitive on the choice of $\gamma$.
	When a spatial bias is re-initialized, we sample a point from $[-1.5, 1.5]\times[-1.5, 1.5]$ pixels uniformly. 
	
	\subsection{Datasets}
	
	\textbf{CIFAR-10/100} datasets \citep{dataset/cifar} consist of 60,000 RGB images of size 32$\times$32 pixels, 50,000 for a training set and 10,000 for a test set. 
	Each image in the two datasets is corresponded to one of 10 and 100 classes, respectively,
	and the number of data is set evenly for each of the classes. We use a standard data-augmentation scheme that is common for this dataset \citep{nips/highway, corr/nin, cvpr/resnet, cvpr/densenet}, 
	i.e.\ random horizontal flip and random translation up to 4 pixels. We also normalize the images in pixel-wise by the mean and the standard deviation calculated from the training set.
	Each model is trained for 300 epochs with mini-batch size 64.
	
	\textbf{Fashion-MNIST} dataset \citep{dataset/fmnist} consists 70,000 gray-scale 28$\times$28 images, 60,000 for a training set and 10,000 for a test set. Each of the labels is associated with one of 10 fashion objects. We use the same data-augmentation scheme with that of CIFAR datasets, along with the dataset normalization scheme. For this dataset, each model is trained for 300 epochs with mini-batch size 128. 
	
	\textbf{Tiny-ImageNet}\footnote{\url{https://tiny-imagenet.herokuapp.com/}} dataset is a subset of ImageNet classification dataset \cite{dataset/ilsvrc}. It consists 200 classes in total, each of which has 500 and 50 images for training and validation respectively. 
	Unlike the ImageNet dataset, each image in this dataset has the spatial resolution of $64\times64$. We also use the data-augmentation scheme of CIFAR datasets, but using random translation of 8 pixels due to the doubled resolution. In order to process the larger images without changing the architecture configurations, we simply doubled the stride of the first convolution in each model.
	In this dataset, each model is trained for 200 epochs with mini-batch size 128. 
	
	\textbf{ImageNet} classification dataset \cite{dataset/ilsvrc} consists of 1.2 million training images and 50,000 validation images, which are labeled with 1,000 classes. 
	For data-augmentation, we perform 224$\times$224 random cropping with random resizing and horizontal flipping to the training images. At test time, on the other hand, 224$\times$224 center cropping is performed after rescailing the images into 256$\times$256. 
	All the images are normalized in pixel-wise by the pre-computed mean and standard deviation.
	Each model is trained for 120 epochs with mini-batch size 128. 
	
	\subsection{Architectures}
	
	\textbf{DenseNet.} 
	In our experiments, we consider four DenseNet \cite{cvpr/densenet} models: DenseNet-40, DenseNet-100, DenseNet-121, and DenseNet-BC-190. 
	DenseNet-40 and DenseNet-100 consist 3 \emph{dense blocks}, each of which consists of $N$ consecutive \emph{dense units}, where each of $N$ is set by 6 and 16, respectively. A dense unit is designed by a 2-layer CNN that produces $k=12$ features, where $k$ is called the \emph{growth rate}. The output of each dense unit is \emph{concatenated} to the input, which defines the main characteristic of the DenseNet architecture. There exist a average pooling layer between the dense blocks for down-sampling, and the final feature maps are pooled into 1$\times$1 features using a global average pooling layer. Unlike \citet{cvpr/densenet}, we do not place a feature compression layer between the dense blocks for simplicity. 
	DenseNet-121 and DenseNet-BC-190 are considered only for the ImageNet experiments and model compression experiments, respectively, and we follow the original architectures specified in \citet{cvpr/densenet}.
	
	\textbf{ResNet and ResNeXt.} We also evaluate our method on pre-activation ResNet-164 \cite{eccv/identity} and ResNeXt-29 ($8\times 64d$) \cite{cvpr/resnext} models designed for CIFAR-10/100 datasets, and ResNet-50 for ImageNet dataset \cite{cvpr/resnet}. Similar to DenseNet, both architectures consists 3 \emph{residual blocks}. We generally follow the model configurations specified by the original papers \cite{cvpr/resnet, eccv/identity, cvpr/resnext}, with some minor modifications for architectural simplicity. Originally, both architectures place a $1\times 1$ convolutional layer of stride 2 between two residual blocks to perform down-sampling and doubling the number of channels. Instead of it, we place a $2\times2$ average pooling layer for down-sampling, and use a simple zero-padding scheme for the doubling.
	
	\textbf{LGC in CondenseNet.} CondenseNet is designed upon DenseNet \cite{cvpr/densenet} architecture, consisting several components to improve the computational efficiency of DenseNet. We primarily focus on the \emph{learned group convolution} (LGC) layer among the components. Architecturally, LGC is a group convolutional layer with $1\times1$ kernel. During training, however, LGC prunes out $\frac{3}{4}$ of its weights through 3 \emph{condensing stages} at $\frac{1}{6}$, $\frac{2}{6}$, and $\frac{3}{6}$ of the total training epochs, $\frac{1}{4}$ for each.
	At each of the condensing stages, channels are scored by the $\ell^1$-norm of the corresponding weights, and the pruning is done according to the scores. Also, LGC adopts channel-wise group-lasso regularization \cite{nips/ssl} during training to induce more sparsity over channels. In our experiment, we consider CondenseNet-182 model. We follow the original setting by \citet{cvpr/condensenet} for training this model and our counterpart CondenseNet-SConv-182 as well. Namely, we train them via SGD with mini-batch size 64 for 600 epochs, and dropout \cite{jmlr/gdropout14} with a drop rate $0.1$ is used. 
	
	\section{Additional Illustrations of De/Re-allocated Channels in DenseNet-40}
	
	\begin{figure}[!h]
		\centering
		\includegraphics[width=0.69\linewidth]{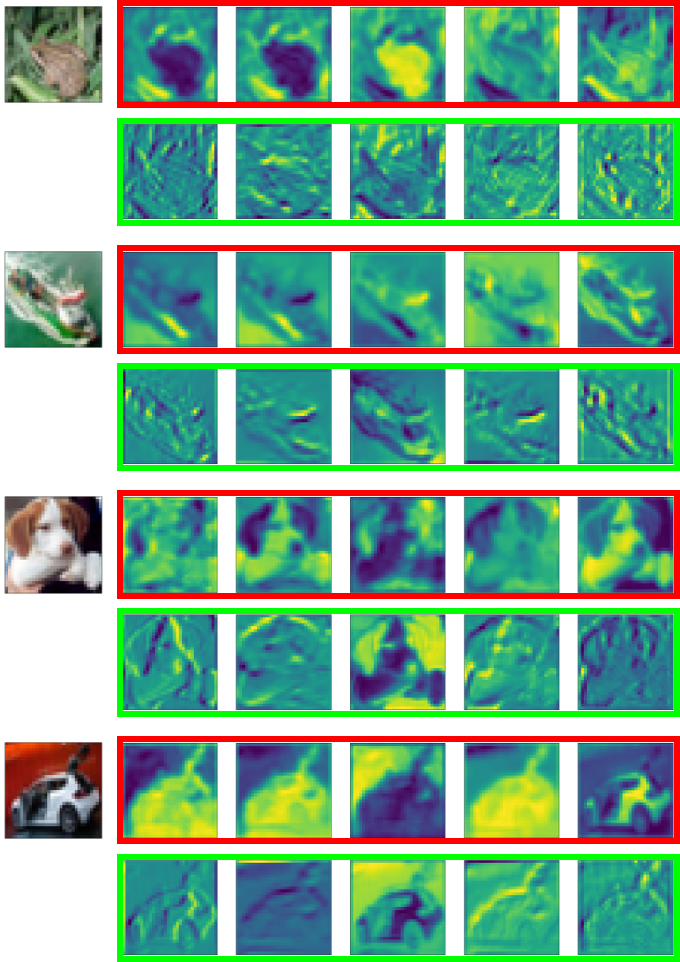}
		\caption{Illustrations of the top-5 feature maps that are most frequently de/re-allocated for each at the $1^{\text{st}}$ dense block of a DenseNet-40 model. The results are shown across four different CIFAR-10 test images. The red and green boxes indicate the top-5 de-allocated and re-allocated channels, respectively.}
		\label{fig:more_examples}
	\end{figure}
	
	\bibliography{references}
	\bibliographystyle{icml2019}